\documentclass[conference]{IEEEtran}
\IEEEoverridecommandlockouts
\usepackage{cite}
\usepackage{amsmath,amssymb,amsfonts}
\usepackage{algorithmic}
\usepackage{graphicx}
\usepackage{textcomp}
\usepackage{subcaption}
\usepackage{wrapfig}
\usepackage{hyperref}
\def\BibTeX{{\rm B\kern-.05em{\sc i\kern-.025em b}\kern-.08em
    T\kern-.1667em\lower.7ex\hbox{E}\kern-.125emX}}
    
\begin{document}
\title{Appearance-based gaze estimation enhanced with synthetic images using deep neural networks}

\author{Dmytro Herashchenko \& Igor Farka\v{s}, \textit{IEEE member} \\ 
\textit{Faculty of Mathematics, Physics and Informatics} \\ 
\textit{Comenius University Bratislava, Slovak Republic} \\
igor.farkas@fmph.uniba.sk
}

\maketitle

\begin{abstract}
Human eye gaze estimation is an important cognitive ingredient for successful human-robot interaction, enabling the robot to read and predict human behavior.
We approach this problem using artificial neural networks and build a modular system estimating gaze from separately cropped eyes, taking advantage of existing well-functioning components for face detection (RetinaFace) and head pose estimation (6DRepNet). 
Our proposed method does not require any special hardware or infrared filters but uses a standard notebook-builtin RGB camera, as often approached with appearance-based methods.
Using the MetaHuman tool, we also generated a large synthetic dataset of more than 57,000 human faces and made it publicly available. The inclusion of this dataset (with eye gaze and head pose information) on top of the standard Columbia Gaze dataset into training the model led to better accuracy with a mean average error below two degrees in eye pitch and yaw directions, which compares favourably to related methods.
We also verified the feasibility of our model by   its preliminary testing in real-world setting using the builtin 4K camera in NICO semi-humanoid robot's eye.
\end{abstract}


\section{Introduction}

Human-Robot Interaction (HRI) is a rapidly growing field with numerous applications in various areas such as manufacturing, healthcare, education, and entertainment \cite{su2023}. One of the most important aspects of HRI is the ability of robots to understand and respond to human behavior and intentions \cite{Thomaz2013}. 
This will make the robots {\it humanized} thank to their ability to recognize human-specific signals \cite{sciutti18}.

Eye gaze is a key signal that people use in social interactions to convey their attention, interest, and emotions. Thus, accurate gaze estimation by robots is an essential human-like ability for successful HRI \cite{Admoni17}.
Several approaches have been proposed for gaze estimation, including model-based, appearance-based, and hybrid methods. However, due to the high variability of gaze directions and head position, as well as lighting conditions in real settings, accurate gaze estimation remains a challenging task.
Therefore, head pose estimation has often been used as a proxy but it is much better observable. Neverheless, it has been argued that direct eye gaze estimation is a better communication cue and the agent's spatial predictor, and cannot be completely substituted by the head pose estimation approach. 
Using an eye tracking system has been shown to allow for a more efficient human-robot collaboration than a comparable head tracking approach, 
according to both quantitative measures and subjective evaluation by the human participants
\cite{palinko2016}.

In this paper, we approach the problem of eye gaze estimation to be used for HRI, not starting from the scratch but instead, taking advantage of existing recent solutions to concrete subproblems such as face detection or head pose estimation toward building a modular system. 
Using existing off-the-shelf components not only saves time and effort but also enables one to narrow down the focus on what is left in solving the problem.
In particular what we add to the system is the neural network module that maps the cropped eyes to 2D gaze information (pitch and yaw).

One of the main differences of this system from others is the ability to work only using a normal RGB camera without any special IR filters and any additional hardware. It also works from different positions relative to the head of the person whose eye gaze it is predicting and should work reliably despite different head positions.

The second contribution of our work is the generated synthetic labeled dataset of 57,375 human faces (using MetaHuman tool) using 15 characters, with controlled head pose and eye gaze directions.\footnote{The dataset is available at \url{http://cogsci.fmph.uniba.sk/metahuman}}. Using this data significantly improved the accuracy of our model compared to only using a smaller training dataset (Columbia Gaze) consisting of real images. 

\section{Related work}

Eye gaze estimation is a research area of great interest due to a high application potential, with numerous different approaches proposed during the last decades. Most of the early methods rely on eye trackers but more recent ones start from processing eye images taken by a camera \cite{Kar2017}.

Traditional approaches to image-based gaze estimation are typically categorized as feature-based or model-based (see, e.g., \cite{Oh2022,Park2018gaze} and references therein).
{\it Feature-based methods} reduce an eye image down to a set of features based on hand-crafted rules and then feed these features into simple machine learning models to regress the final gaze estimate. 
They generally have low requirements for structured data and computational power. Nevertheless, they often work best with IR cameras and lighting where the pupil becomes very bright and easy to distinguish, and become a lot less accurate when using normal RGB cameras, especially in different lighting conditions.

{\it Model-based methods} instead attempt to fit a known 3D model to the eye image by minimizing a suitable loss function.
Effective use of model-based machine learning methods can provide significant benefits. However, this approach requires a significant amount of labeled data to train the models effectively and can be computationally demanding to implement.

A more recent category of the so-called {\it appearance-based models} became popular, enabled by the rise of deep networks, especially convolutional neural networks. These are actually end-to-end methods applied within supervised learning assuming that labeled data are available.
Early works in appearance-based methods were restricted to laboratory settings with a fixed head pose. These initial constraints have become progressively relaxed, notably by the introduction of new datasets collected in everyday settings or in simulated environments.
Several CNN architectures have been proposed for person-independent gaze
estimation in unconstrained settings, mostly differing in terms of possible input data modalities \cite{zhang15_cvpr}. 
As argued in \cite{Kar2017}, direct, consistent comparisons of methods are difficult due to lacking standardized methodologies but approximately we can say that appearance-based methods, according to Table II in \cite{Kar2017}, do not reach accuracy below $3^{\circ}$ in case of models with free head movement.

In our work we tackle the problem of a real-world eye gaze estimation with changing lighting conditions, head position, and noisy images that make it more difficult and unreliable. 

\section{Datasets for eye gaze estimation}
\label{sec:datasets}

There exists a variety of eye gaze datasets which are both real-life and synthetic. In this section, we will explore some of them, mentioning their pros and cons. In addition, we will briefly describe the process of generating our own dataset using the MetaHuman tool aiming to improve the accuracy of our eye gaze estimation model.

\subsection{Columbia Gaze Data Set}

One of the most diverse datasets that were taken in a controlled environment and are using degrees to represent pitch and yaw is the Columbia gaze dataset \cite{smith2013}. The 5,880 photos of 56 people provide a comprehensive set of eye gaze data with different gaze directions and head poses. In terms of the number of people, this dataset is superior to other eye gaze datasets that were publicly accessible at the time of its release. The individuals in the sample come from various ethnic backgrounds and as an important feature, almost half of them wear glasses.

\begin{figure}[ht]
  \begin{center}
\includegraphics[width=0.48\textwidth]{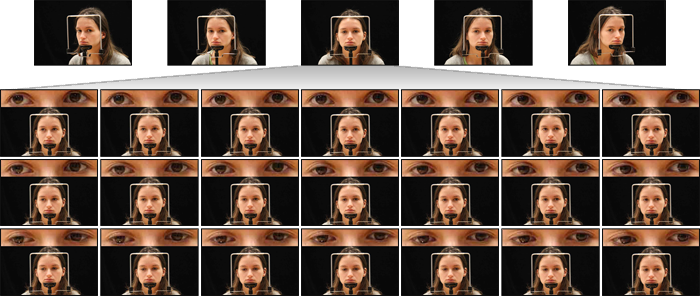}
  \caption{A sample of Columbia gaze dataset \cite{smith2013}.}\label{fig:columbia}
  \end{center}
\end{figure}

Each high-quality image in this dataset has a resolution of 5,184$\times$3,456 pixels, which is great for predicting the dataset itself, but is not representative of the average camera used for these purposes, so it does not help with the training model to better predict data in varying conditions. Another problem is that the lighting conditions are almost constant across the dataset, which hinders the robustness of the trained models. 
In Figure~\ref{fig:columbia} we can see a sample from the dataset with different eye and head positions.

We chose Columbia as the primary dataset because of its variety of people and and their head positions. For comparison, MPIIGaze dataset \cite{zhang15_cvpr} contains almost 214,000 images with 15 characters, but these are much less varied in terms of head and eye positions. 
The Unity eye dataset \cite{wood2016_etra} also provides the Unity eyes tool to generate different images of eyes from a high-resolution 3D model and one million images generated with the tool as a dataset. However, it lacks the possibility of adding accessories such as glasses, it does not allow control over the lighting conditions of the scene and does not provide images of the full face since it does not exist. 

\begin{figure}[t!]
\centering
\includegraphics[width=0.49\textwidth]{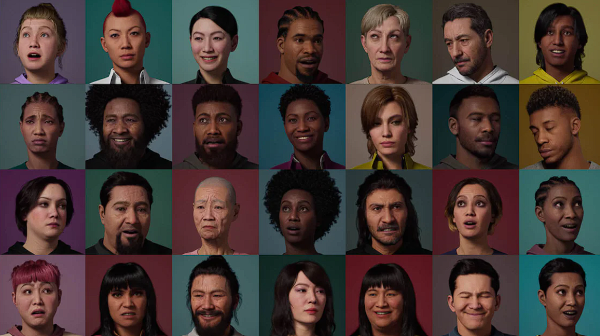}
\caption{Synthetic faces generated by the MetaHuman tool \cite{games2021}.}\label{fig:metahumans}
\end{figure} 

\subsection{MetaHuman dataset}
\label{sec:metahuman-dataset}

In order to expand the training data set, we took advantage of existing generative AI tools and synthesized human faces in a perfectly controlled way. MetaHuman tool for Unreal Engine is an innovative technology enabling the creation of highly realistic and flexible digital human models. These in combination with the popular game engine produce highly realistic and immersive experiences exploitable not only in traditional computer games but mainly in virtual reality, augmented reality, and even movie production.  
Using this tool, we generated a dataset of 57,375 images using 15 characters in different eye and head positions. 
Examples of these realistic humans can be seen in Figure~\ref{fig:metahumans}.

\begin{figure}[t!]
\centering
\includegraphics[width=0.48\textwidth]{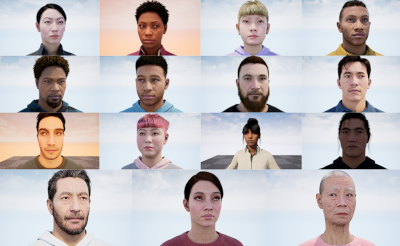}
\caption{A sample of images from our generated dataset.}
\label{fig:mh-dataset}
\end{figure}

In the image generation process, we proceeded as follows:
The existing database contains a large of virtual humans, so we selected a diverse set of 15 characters for further processing.\footnote{It would be possible, but not trivial, to add attributes such as glasses to the characters, since one would need a third-party 3D modeling software and basic modeling skills.}
The next step was to set up the scene, where one can change lighting conditions to make the dataset more compatible with real-world case and at the same time more robust for training. 
Finally, we generated a list of different values for both head and eye positions and iterated over them while taking snapshots using the animation controller. This way, each character went through 153 eye positions in each of the 25 head positions, resulting in 3825 images generated per character. Figure~\ref{fig:mh-dataset} provides examples of generated images.

\section{The eye gaze estimation system}\label{sec:model}

\begin{figure}[ht]
\centering
\includegraphics[width=0.5\textwidth]{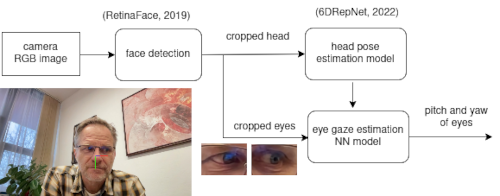}
\caption{Architecture of our eye gaze estimation system. Cropped eyes are enlarged to be better visible.}
\label{fig:overall-arch}
\end{figure}

Our proposed eye gaze estimation system builds on exiting modules and adds to these the neural network module. All components are described in mode detail below. The overall layout of the architecture is shown in Figure~\ref{fig:overall-arch}. 

An input image is taken from  
RGB camera using OpenCV and is rescaled to match input sizes for the models used. RGB values are scaled to the interval [0,1]. For model building we used existing images (from both datasets), in real we tested the system using the notebook's mounted camera and also the 4K camera in NICO robot's vision. Now we describe indvidual modules of the system.

\subsubsection{Face detection}
\begin{figure}[t!]
\centering
\includegraphics[width=0.48\textwidth]{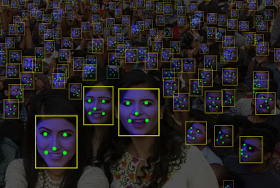}
 \caption{Cropped image using the RetinaFace module \cite{deng2019}.}\label{fig:retinaface}
\end{figure}

We used a well-known RetinaFace \cite{deng2019}, the pre-trained model that works well on a variety of datasets. It provides accurate positions of all faces on the image at multiple spatial scales. It also finds five important locations on the faces, corresponding to the mouth, nose and the eyes, which we used for eye cropping.
We used the implementation from \cite{RetinaGit} GitHub repository. The performance of RetinaFace is illustrated in Figure~\ref{fig:retinaface}.

\begin{wrapfigure}{r}{0.25\textwidth}
\centering
\includegraphics[width=0.25\textwidth]{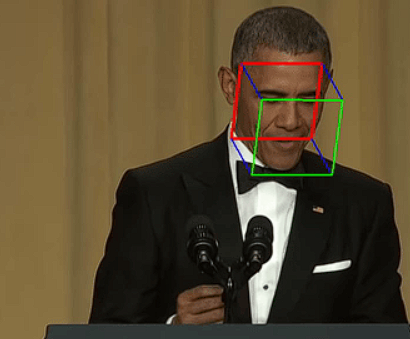}
\caption{Head pose estimation using the 6DRepNet module \cite{Hempel22}.}\label{fig:6drepnet}
\end{wrapfigure} 

\subsubsection{Head pose estimation}
After face segmentation (only one needed in our case), we apply to the cropped image the 6DRepNet \cite{Hempel22}, a state-of-the-art pre-trained model for head pose estimation. It ranks very well on a couple of popular head pose datasets and reliably gets to the top 5 on others (as of writing this paper). 
The ability of 6DRepNet to learn and predict head positions beyond the narrow-angle constraints of most other models is one of its key advantages. The proposed network breaks this limitation by capturing the entire rotational view of the head, whereas previous approaches limited head pose estimation to a short range of angles to produce good results.
In our case, head pose information was used for providing additional data to our main eye gaze estimation model for better accuracy (especially if eye are well visible). Figure~\ref{fig:6drepnet} illustrates its function.

\subsubsection{Eye gaze estimation} The final step involves estimating the eye gaze using a custom-built convolutional neural network (CNN) model. It takes as inputs two cropped eyes that are then concatenated together as shown in Figure~\ref{fig:overall-arch} as the input, and outputs two values in degrees. These are the predicted pitch and yaw of the eyes. 

\begin{figure}[ht]
  \centering
\includegraphics[width=0.45\textwidth]{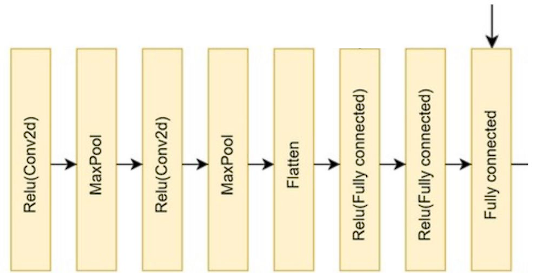}
\caption{Architecture of the eye gaze estimation CNN that outputs 2D prediction signal (pitch and yaw in degrees).}
\label{fig:cnn-architecture}
\end{figure}

We experimented with an CNN architecture and the resulitng version is shown in Figure~\ref{fig:cnn-architecture}. To meet the unique needs and difficulties of better eye gaze estimation in the context of human-robot interaction, this architecture has undergone a number of enhancements and fine tuning.

We will provide more details about the model in the next section. Convolutional layers are expectedly responsible for learning different features (e.g. pupils, eyelids, etc.) and these are then processed by fully connected layers to yield the desired 2D eye gaze information (in degrees). The estimated head position is provided to the last fully connected layer with an expectation to contribute to calculating the final result.

\section{Results}

We provide results obtained with implemented, trained and tested models, also comparing the effects of the both datasets (Columbia Gaze dataset and MetaHuman dataset). In doing so, we can evaluate the effectiveness of generating a dataset in real-world scenarios and evaluate its capacity to replace real datasets that are much harder and more expensive to produce. We will also be able to pick the best model to use with a robot for real-life human-robot interaction.

A regular notebook with built-in GPU was used for both training and inference at first, although later on we started to use a more powerful PC for faster training times. But the inference could be run on a notebook with around 30 FPS.

\subsection{CNN model building}

The accuracy and robustness of the eye gaze estimation system were improved step by step by various ways of hyperparameter tuning and by model architecture to achieve the best results. To evaluate the CNN model, we used a four-fold cross-validation on the Columbia Gaze dataset. For training we minimized the mean squared error and for testing the model we used the mean absolute error (MAE), to be able to compare results with other models. All the error values are averaged for both output values (pitch and yaw). 

The model building process involved several key steps most of which contributed to a smaller or larger improvement in performance. 
Originally, the crop size for eye images was set to 300$\times$900 pixels, which yielded an average error rate MAE = $3.4^{\circ}$. Reducing the crop size to  100$\times$300 resulted in a significant reduction in model training (2.5 times) and inference time while preserving the accuracy. 
At the same time, it sped up the process of searching for an optimal model architecture.

As for real-world testing, using a lower-resolution webcam was found beneficial. Also, the reason for using two separate concatenated images rather than one joint image (cropping both eyes together) was that our method allows for segmenting narrower rectangles for each eye, especially for rolled head poses, hence reducing the dimensionality of both images.

Among various optimization algorithms we tried, Adam was found best \cite{Kingma17} and was hence used in subsequent simulations.

We also experimented with appropriately integrating the head pose information estimated by 6DRepNet into the model. Adding this information to deeper layers did not help, perhaps due to huge number of feedforward channels coming from convolutional layers. However, adding head pose information to the last hidden layer, containing only 53 units, had a positive effect and led to improvement in the model accuracy approximately by $0.2^{\circ}$. Also empirically, we observed that the model appeared more robust when the subject was turning the head in different directions.

The last improvement to the MAE we were able to get was from expanding the training data by image mirroring, which is not really model optimization but is still important for model improvement. This helps because our faces are not symmetrical, and the lighting conditions are not perfect either, so the flipped images are a little different. This way, we were able to double the dataset size, including the cropped eyes and improve our models' performance to MAE $\approx 1.9^{\circ}$. The maximum error did not change, though.

After all the improvements described above, we arrived at a final architecture shown in Figure~\ref{tab:network_architecture} having 2,092,342 trainable parameters.

\begin{table}[ht]
\centering
\caption{Summary of the network architecture}
\begin{tabular}{llr}
\hline
Layer (type) & Output Shape & No. of parameters \\
\hline
Conv2d-1 & [9, 68, 208] & 252 \\
MaxPool2d-2 & [9, 22, 69] & 0 \\
Conv2d-3 & [26, 20, 67] & 2,132 \\
MaxPool2d-4 & [26, 6, 22] & 0 \\
Linear-5 & [600] & 2,059,800 \\
Linear-6 & [53] & 30,050 \\
Linear-7 & [2] & 108 \\
\hline
\end{tabular}
\label{tab:network_architecture}
\end{table}

\subsection{Dataset testing and comparison}
\label{sec:dataset-testing}

As a next step, we investigated the effect of the used data set(s) on model accuracy. We compared three scenarios: Columbia dataset, MetaHuman dataset, and a combination of both. To avoid overfitting, we applied early stopping using the validation subset.
After running 4-fold cross-validation across Columbia Gaze dataset, MetaHuman dataset and the combined dataset, we achieved results shown in Table~\ref{tab:crossval}.

\begin{figure}[t!]
\centering
\includegraphics[width=0.48\textwidth]{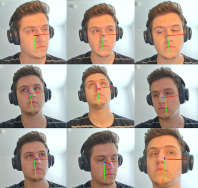}
  \caption{Example images from the real-world testing of our system using a notebook built-in camera. Blue line points to the predicted eye gaze direction.}
  \label{fig:cam-test}
\end{figure} 

\begin{table}[h]
\centering
\caption{Cross-dataset results of eye gaze estimation, in terms of the mean and standard deviation.}
\begin{tabular}{l|rrr}
\hline
Trained / Tested & Columbia Gaze & MetaHuman & Combined \\
\hline
Columbia Gaze & 1.93$\pm$1.50 & 8.42$\pm$4.83 & 7.55$\pm$5.17 \\
MetaHuman     & 8.12$\pm$5.81 & 0.59$\pm$0.52 & 1.90$\pm$3.79\\
Combined      & 2.88$\pm$1.94 & 0.65$\pm$0.56 & 1.04$\pm$1.25 \\
\hline
\end{tabular}
\label{tab:crossval}
\end{table}

As can be seen, the most stable model across all three datasets is the one trained on the combined dataset, as expected, which means is that by combining the two datasets we get the most diverse and generalized model. This model is the one we have chosen to be the main model.

If we look at the model only trained on the Columbia Gaze Dataset, we can compare it to other models that were trained in other papers. For example, one of the best models presented in \cite{Park2018gaze} achieves MAE $\approx 3.8^{\circ}$ on the Columbia Gaze Dataset with 5-fold cross-validation, whereas our model only MAE $\approx 1.93^{\circ}$. We have to acknowledge, though, that our model has a problem with stability in worse lighting conditions, and that is exactly why the model trained on the combined dataset is superior.
Interestingly, the model trained and tested on MetaHuman data performs best, and cross-wise, both models, when trained on one and tested on the other dataset, perform worst, yielding MAE$\approx8.0^{\circ}$.

\subsection{Real world testing}
\label{sec:real-world}

We performed preliminary tests of our model beyond  the given datasets, using two methods. 
First, we tested the system (trained in all three scenarios mentioned in Table~\ref{tab:crossval}) in real time on ourselves.
Figure~\ref{fig:cam-test} shows example photos (of the first author) testing the model using different head positions and light conditions. We observed that the model trained on the combined dataset performs better in worse lighting scenarios than the one trained only on the Columbia Gaze dataset.

\begin{wrapfigure}{r}{0.2\textwidth}
\centering 
\vspace{-3mm}
\includegraphics[angle=270,width=0.20\textwidth]{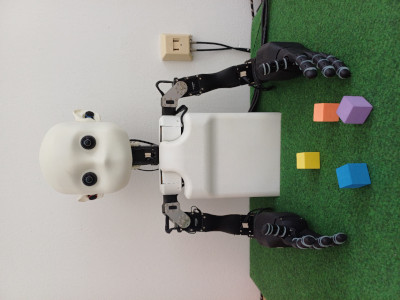}
\caption{Semi-humanoid NICO robot has an 4K RGB camera in each eye.}
\vspace{-3mm}
\label{fig:nico}
\end{wrapfigure} 

\vspace{1cm}

\begin{figure}[t!]
\centering 
\includegraphics[width=0.48\textwidth]{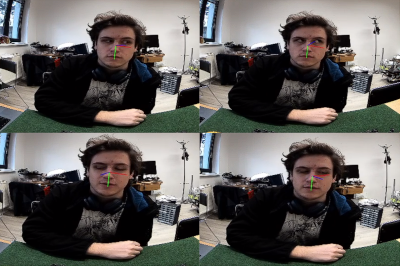}
   \caption{Example images from the real-world testing with the NICO robot.}
   \label{fig:nico-test}
\end{figure}

Second, we tested the eye gaze estimation model with the 4K camera built in NICO robot \cite{nico2017} (Figure~\ref{fig:nico}), a platform created to study human-robot interaction. 
Examples images of the model predicting eye direction are provided in Figure~\ref{fig:nico-test}.
Despite high resolution of NICO's (fish eye) camera the model performs well (thanks to reliable RefinaFace module). Nevertheless, we think that performance can further be improved even by calibrating the lens correcting algorithm on the camera. This would make the geometry on the pictures closer to human perception and probably make the gaze estimation process easier. 

Also, after testing our model that was only trained on the MetaHuman dataset in the real world, we found it can successfully predict the eye gaze of real people, so we think that generating datasets with modern game engine technologies like Unreal Engine has a promising future. One reason is that it allows creating a dataset of visual or 3D data for a wide variety of tasks at significantly lower costs and time compared to doing so in the real world. We provide the code and the pre-trained model on   GitHub\footnote{\url{https://github.com/flakeua/BachelorsThesis}}. 

We plan to use the model in our HRI research on intention reading by a robot in the context of objects placed at the table. To predict the focus of human attention, the predicted eye gaze will be extrapolated towards the table and the best matching object on the table will be chosen. Although we have not yet tested it, we believe that achieved accuracy (below $2^{\circ}$) will be sufficient for disambiguating the target object.

\section{Discussion}

In this work, we dealt with a human-oriented task of robust eye gaze estimation, which is an important component for the human-robot interaction. Eye gaze serves as in important overt cue that can enable the robot to predict human behavior and possibly human intentions. We took advantage of existing well functioning modules (RetinaFace and 6DRepNet) that allowed us to narrow down our focus on a subproblem of learning the mapping (regression) from the cropped eyes (in combination with the head pose) to gaze direction (in terms of pitch and yaw).

Deep convolutional neural networks are well-known nonlinear function approximators that are suited well for such mapping tasks. 
The flexibility of our suggested solution, to operate with a standard RGB cameras without the need for additional hardware or IR filters, is one of its distinctive features. The system was created to function effectively in varying head and eye positions, with the camera position independent of the head position.

Using the potential of Unreal Engine and MetaHumans tools in terms of creating a much larger dataset was beneficial for our research but the availability of the dataset is likely to become a very useful resource for further research in this area. There is also a potential to leverage the cutting-edge capabilities of Unreal Engine to generate more complicated data sets that are more diverse, have harder lighting scenarios, and have characters wearing glasses, possibly creating opportunities for the creation of more sophisticated and precise eye gaze estimation models.

Despite encouraging results of our system, we intend to improve its robustness by training it on more variable light conditions, and people wearing glasses. Another challenge to be tackled (by any system using a standard RGB camera) is the prediction of eye gaze directions when a person is looking down when the eye pupils are not well visible. The easiest, technical way to fix this problem would be to place another camera at a lower position. Alternative, human-like solution would be to make the model rely more on the head pose in such cases, or take advantage of the context knowledge that navigates the target prediction.

Also we should mention that we are aware of two potential difficulties that we will need to tackle in forthcoming research. One is that the robot will be moving its head (looking directly at a human or at a table) during interaction, so the extrapolation will have to include that transformation. The other thing is that the participant will use a head-mounted eye-tracker PupilCore in order to record his/her eye gaze data, which may worsen the model accuracy.

Overall, the motivation for this work is strong. 
Integrating gaze estimation systems into HRI has far-reaching practical implications, enhancing communication, safety, personalization, and overall user experience. 
Human gaze tracking enhances robot behavior by providing critical context and communication cues. It enables robots to be more attentive, adaptive, and responsive to human needs and preferences, ultimately leading to more effective and engaging human-robot interactions.
Hence, it will support a win-win situation, by enabling the robot to get engaged in a smoother and more trustworthy interaction that will be more appreciated by the human user, even without language.

\section*{Acknowledgment}
This research was supported by the national project APVV-21-0105 and by the Horizon Europe project TERAIS, grant number 101079338.

\bibliographystyle{IEEEtran}
\bibliography{gaze-main}

\end{document}